\documentclass[letterpaper]{article}
\usepackage{aaai}
\usepackage{hyperref}
\usepackage{times}
\usepackage{helvet}
\usepackage{courier}
\usepackage{graphicx}
\usepackage{times}
\usepackage{algorithmic}
\usepackage{latexsym}
\usepackage{booktabs}
\usepackage{amsmath}
\usepackage{amssymb}
\usepackage{bm}
\usepackage{natbib}
\usepackage{CJKutf8}
\usepackage{url}
\usepackage{tabularx}
\usepackage{subfig}
\usepackage{multirow}

\usepackage{color}
\usepackage[ruled,linesnumbered]{algorithm2e}
\begin{document}
%
\title{Extracting Relational Triples Based on Graph Recursive Neural Network via Dynamic Feedback Forest Algorithm}
\author{
Hongyin Zhu\\
hongyin\_zhu@163.com\\
}
\maketitle
\begin{CJK*}{UTF8}{gbsn}
\begin{abstract}
Extracting relational triples (subject, predicate, object) from text enables the transformation of unstructured text data into structured knowledge. The named entity recognition (NER) and the relation extraction (RE) are two foundational subtasks in this knowledge generation pipeline. The integration of subtasks poses a considerable challenge due to their disparate nature. This paper presents a novel approach that converts the triple extraction task into a graph labeling problem, capitalizing on the structural information of dependency parsing and graph recursive neural networks (GRNNs). To integrate subtasks, this paper proposes a dynamic feedback forest algorithm that connects the representations of subtasks by inference operations during model training. Experimental results demonstrate the effectiveness of the proposed method.
\end{abstract}

\section{Introduction}
\begin{figure}[!ht]
\centering
\includegraphics[width=2.5in]{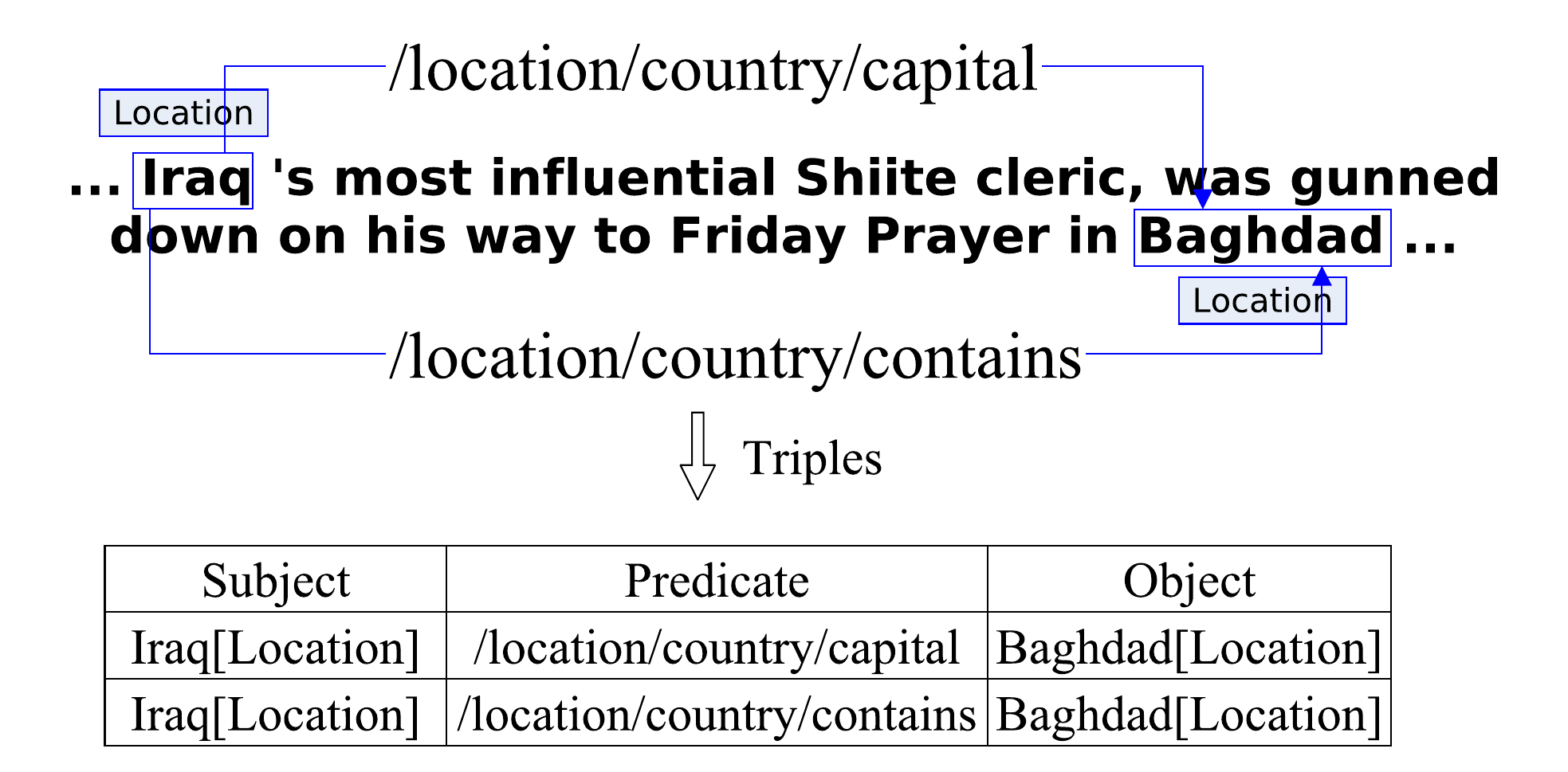}
\caption{An example to extracting the triples from unstructured text}
\label{example}
\end{figure}

Extracting triples from unstructured documents is a fundamental technology in constructing knowledge bases. This process involves identifying and extracting (subject, predicate, object) from textual input, as depicted in Figure \ref{example}. Named entity recognition (NER) and relation extraction (RE) methods can collaborate in the triple extraction pipeline to identify entities and relations, respectively.

Prior works mainly train these two subtasks separately because there is a cascade process between them, and the output of the NER and the input of the RE subtask are completely different. Despite the flexibility of the pipeline methods, they block the connection between two subtasks. Some approaches [\cite{miwa2016end}] overcome the joint extraction problem by sharing features of two subtasks, but two-stage models require separate model training and it is difficult to fully exploit the relationship between subtasks. Despite task conversion resolving this issue by eliminating the cascade process, i.e., converting joint extraction into the sequence labeling process \cite{zheng2017joint}, their performance is potentially limited by the design of the new tag scheme. Ideally, we hope to overcome these two problems, by integrating two subtasks in one model and not modifying the original tag scheme. 

Smoothly integrating cascaded tasks is a nontrivial task. The difficulty lies in that the relation classification (RC) aims to classify the relation between an entity pair while we have not known where are the entities. The main gap between NER and RC subtasks is how to generate relation candidates, which makes model integration a challenging issue. This paper introduces a dynamic feedback forest algorithm (DFF) to integrate subtasks into a joint optimization process by using inference operations during model training. These operations can directly connect the representations of NER and RE. Then, the model can use the original tag to supervise the two subtasks together. The DFF algorithm dynamically constructs multiple subgraphs to form the forest and teach the model according to the deviation of current prediction and ground truth. This algorithm executes the inference and feedback operations during model training, adapting the model to new samples. 

The second challenge is to unify the representations of entity and relation. This paper maps the entity and its relation to the dependency graph. We applied the idea of the recursive neural network to the dependency graph so that we could directly label a vertex as an entity, a relation instance, or none. Sometimes, there are different relations between the same entity pair, i.e., ``{\it Iraq's capital is Baghdad}" and ``{\it Iraq contains Baghdad}", as shown in Figure \ref{example}. The prior new tag scheme is incapable of dealing with overlapping relations (an entity belongs to more than one relation). Other works [\cite{wang2018joint,zeng2018extracting}] design new strategies to solve this problem. Our GRNN model is naturally compatible with different relations between an entity pair without changing the network. To further simplify the network, we also test the hypothesis that a unified representation can be obtained to resolve two subtasks by our GRNN.

We conduct experiments on the NYT\footnote{\url{https://github.com/shanzhenren/CoType}} dataset. Experimental results demonstrate the effectiveness of our approach. The major contributions of this paper can be summarized below.

(i) This paper explores a graph labeling approach to resolving the triple extraction task.

(ii) This paper utilizes graph recursive networks to model the triple extraction task with joint task learning and joint representation learning.

(iii) This paper introduces the DFF algorithm, which enables the model to connect representations of different subtasks through inference operations during training, facilitating the learning process.

\section{Related Work}

For the joint extraction task, \cite{miwa2016end} propose a method using sequential and separate tree-structured LSTM-RNNs for NER and RC. Their work represents the relation by the shortest dependency path, while our model uses a vertex to represent any entity or relation. Their method use scheduled sampling, entity pretraining, and label embedding to enhance the training process, while we use the maximum log-likelihood with the DFF algorithm. 

\cite{ren2017cotype} use a domain-agnostic segmentation algorithm to mine entity mentions, and convert the task into a global embedding problem. 
\cite{khashabi2013recursive} present a recursive neural network-based method to extract entity and relation separately, but leave the joint learning for future work. Our model solves the joint learning problem. 

\cite{zheng2017joint} propose a novel tag scheme to convert this task into a sequence labeling problem, but it increases about eight times of classes and cannot deal with the overlapping relations. \cite{wang2018joint} propose a neural transition-based approach and a suite of transition schemes, while our method is only based on the graph structure of dependency parsing. \cite{zeng2018extracting} propose the One-Decoder and Multi-Decoder approaches to extract relational facts with copy mechanism. They divide the sentences into three types according to the triplet overlap degree, and our approach is also compatible with overlapping relations.
\cite{liu2018seq2rdf} propose the Seq2RDF which aims to map textual input to existing RDF triples, so it relies on the knowledge graph vocabulary, while our approach aims to extract triples directly from unstructured documents in a general way.

\section{Methods}
\label{methods}
\subsection{Task Definition}
Given a text sequence $x=[w_1,...,w_n]$ where $w_i$ is the $i$-th token, the triple extraction task aims to extract multiple ($s_{[type]}$, $p$, $o_{[type]}$) where $s$ and $o$ represent two non-overlapping consecutive spans, $s=[w_{s1},w_{s1+1}...,w_{s2}]$ and $o=[w_{o1},w_{o1+1}...,w_{o2}]$. $p \in R$ is the relation type, where $R$ is a set of predefined relation types. The subscript $_{[type]} \in C$ is the entity type, where $C$ is a set of predefined entity types. 

\subsection{The Graph Labeling Scheme}
\begin{figure}[h]
\centering
\includegraphics[width=2.8in]{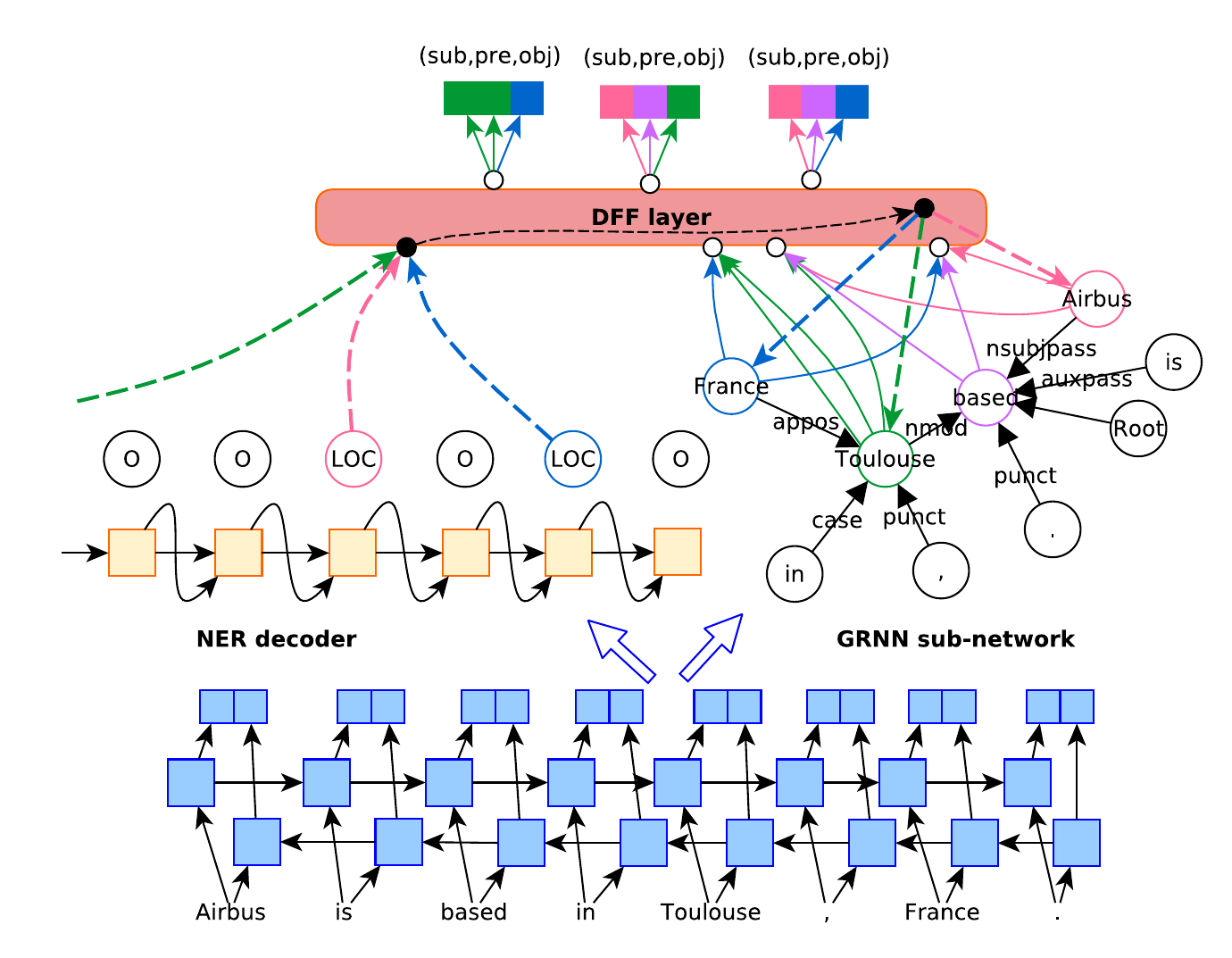}
\caption{Joint task learning (JTL) network for integrating subtasks}
\label{figarch}
\end{figure}

Figure \ref{figarch} uses an example to demonstrate the idea of how to convert the joint extraction task into a graph labeling problem. The lower layer uses a Bi-LSTM encoder to generate the contextual representation. Then, the intermediate layer uses the Stanford CoreNLP [\cite{manning2014stanford}] to convert the sentences into dependency graphs. This model uses vertices to represent entities and relations. Finally, this model integrates the subtasks using the DFF algorithm (introduced in subsection \nameref{dff}).

\begin{figure}[h]
\centering
\includegraphics[width=1.8in]{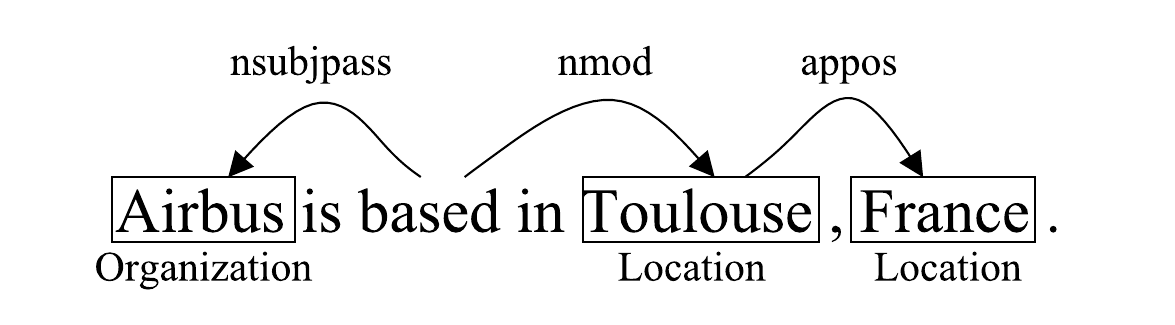}
\caption{An example to demonstrate the entity and relation representation}
\label{sentence}
\end{figure}

We take a simple example to demonstrate the graph labeling scheme. Figure \ref{sentence} contains three entities (Airbus [ORG], Toulouse [LOC], and France [LOC]). Each entity is mapped to the corresponding vertex. Each relation is mapped to the least common ancestor (LCA) of two entity subgraphs, i.e. the relation of (Airbus, relation, Toulouse) is represented by the vertex ``based'' which is the LCA of ``Airbus'' and ``Toulouse'', which implies the relation type is ``{\it /business/company/place\_founded }".

\begin{figure}[h]
\centering
\includegraphics[width=3in]{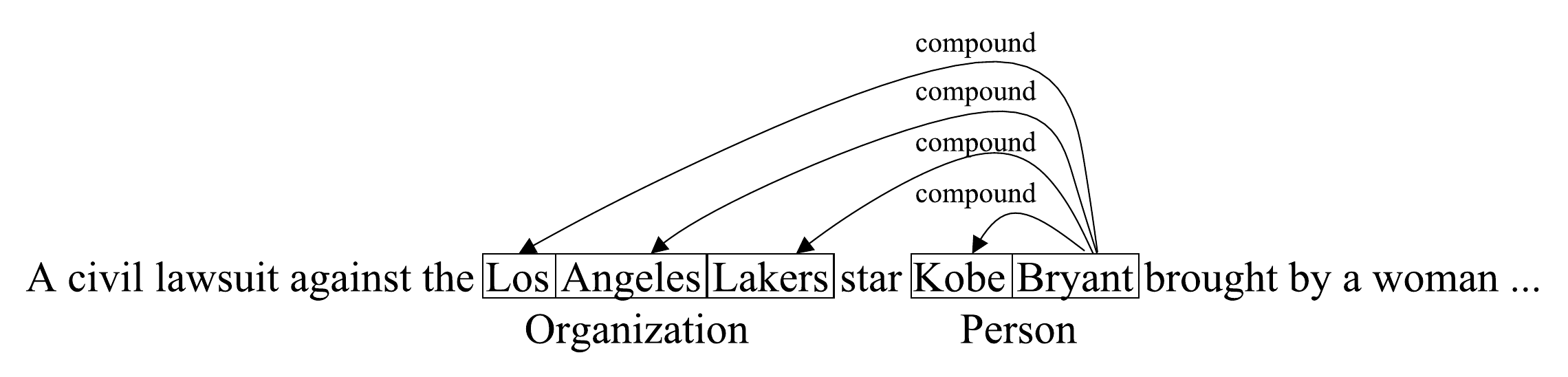}
\caption{An example that two entities share the same root}
\label{example2}
\end{figure}

Any relation or entity can be represented by a vertex, so this model can directly classify the vertices. For the entity phrase, this model maps each entity as a subgraph and takes the root vertex as the representation like [\cite{khashabi2013recursive}]. 
In some cases, two entities share the same root, for example, the {\it Los Angeles Lakers} and {\it Kobe Bryant} of Figure \ref{example2} share the same root {\it Bryant} in the dependency graph. To better differentiate the entities, an entity vector is finally composed of the vertex representation and the average pooling of the entity phrase. 

Sometimes, there are different relations between two entities. In addition to the 1-of-n classification, this model can be extended to the multi-label classification form which can extract different relations of the same entity pair by multi-label graph labeling for each vertex.

\subsection{GRNN Units}
We have obtained the graph structure of each sample. The gated units and CNN have achieved impressive performances in many deep-learning models. To better process the graph data, we implement different neural units (LSTM [\cite{hochreiter1997long}], GRU [\cite{chung2014empirical}] and CNN [\cite{lecun1998gradient}]) that are compatible with our GRNN model. 

{\bf Graph RNN}
For the basic perceptron, each source vertex accumulates the information from its target vertices through a non-linear activation function as below.
\begin{align}
\label{grnn}
h_t=tanh(Wx_t + \Sigma_{j=0}^{P(t)}U_{m(t,j)} h_j +b)
\end{align}
where $W \in \mathbb{R}^{d\times l},U \in \mathbb{R}^{m \times d \times d}, h \in \mathbb{R}^{d},x \in \mathbb{R}^{l},b \in \mathbb{R}^{d}$. The $t$ and $j$ are the indexes of the source and target vertices respectively. $P(t)$ is the number of target vertices of the source vertex $t$. $U$ is a group of edge parameters where $m$ denotes the 193 edge types, and $d \times d$ is the dimension of an edge embedding. $U_{m(t,j)}$ denotes the edge embedding that connects source $t$ and target $j$. Considering edge types can make the model control substreams in different weights. $h$ and $x$ denote the vertex representation and input respectively. To resolve the gradient vanishing problem, we also implemented the gated units.

{\bf Graph LSTM}
For the LSTM unit, this paper simplifies the network by using a group of edge parameters (same as the variable $U$ in formulation \eqref{grnn}) based on the study of [\cite{peng2017cross}]. Compared with the linear chain LSTM unit, the main improvement is the separate forget gate for each input edge, which can achieve selective control of different edges. 

\begin{align}
\label{lstm}
i_t&=\sigma(W_i x_t +\Sigma_{j=1}^{P(t)} U_{m(t,j)} h_j +b_i) \\
f_{m(t,j)}&=\sigma(W_f x_t + U_{m(t,j)}h_j + b_f ) \\
o_t&=\sigma(W_o x_t + \Sigma_{j=1}^{P(t)} U_{m(t,j)} h_j + b_o) \\
\tilde{c_t}&=\tanh(W_c x_t +\Sigma_{j=1}^{P(t)} U_{m(t,j)} h_j +b_c) \\
c_t&=i_t \odot \tilde{c_t} + \Sigma_{j=1}^{P(t)} f_{m(t,j)} \odot c_j  \\
\label{cell}
h_t&=o_t \odot \tanh(c_t)
\end{align}
where $h$, $c$, and $o$ are the hidden state, the cell state, and the output respectively. $W$, $U$ and $b$ are model parameters. The $\odot$ represents the Hadamard product (pointwise multiplication).

{\bf Graph GRU} Analogy with the graph LSTM unit, the graph GRU unit separates the reset gate for each edge. In the graph LSTM, the output gate, and the cell state can limit the hidden state into an effective range as formulation \eqref{cell}. However, for the large values of hidden states, the outputs of graph GRU might grow large in magnitude. 
To counteract this effect, this paper adds a non-linear activation in the edge to limit the response in practice.
\begin{align}
\label{gru}
p_{j} &=\tanh(U_{m(t,j)} h_{j}) \\
z_t&=\sigma(W_z x_t+\Sigma_{j=1}^{ P(t)} p_{j}+b_z) \\
r_{m(t,j)}&=\sigma(W_r x_t +b_r+p_{j}) \\
\label{acc}
\tilde{h_t}&= \sigma(W_h x_t+ U_r\Sigma_{j=1}^{ P(t)}(r_{m(t,j)} \odot p_{j})+b_h)\\
\label{grud}
h_t&=(1-z_t)\odot \Sigma_{j=1}^{ P(t)} p_{j} + z_t \tilde{h_t} 
\end{align}
where $h$, $z$, and $r$ are the output vector, update gate vector and reset gate vector. $W$, $U$ and $b$ are model parameters.

{\bf Graph RCNN} The recursive neural network [\cite{socher2013parsing}] can only process the binary combination and is not suitable for graph data, since a source vertex may have two or more targets. This paper adopts the RCNN unit [\cite{zhu2015re}] which can deal with the k-ary parsing tree. To make the RCNN unit compatible with our GRNN, we add different edge types and generalize the RCNN unit to DAG.

\begin{figure}[h]
\centering
\includegraphics[width=2.8in]{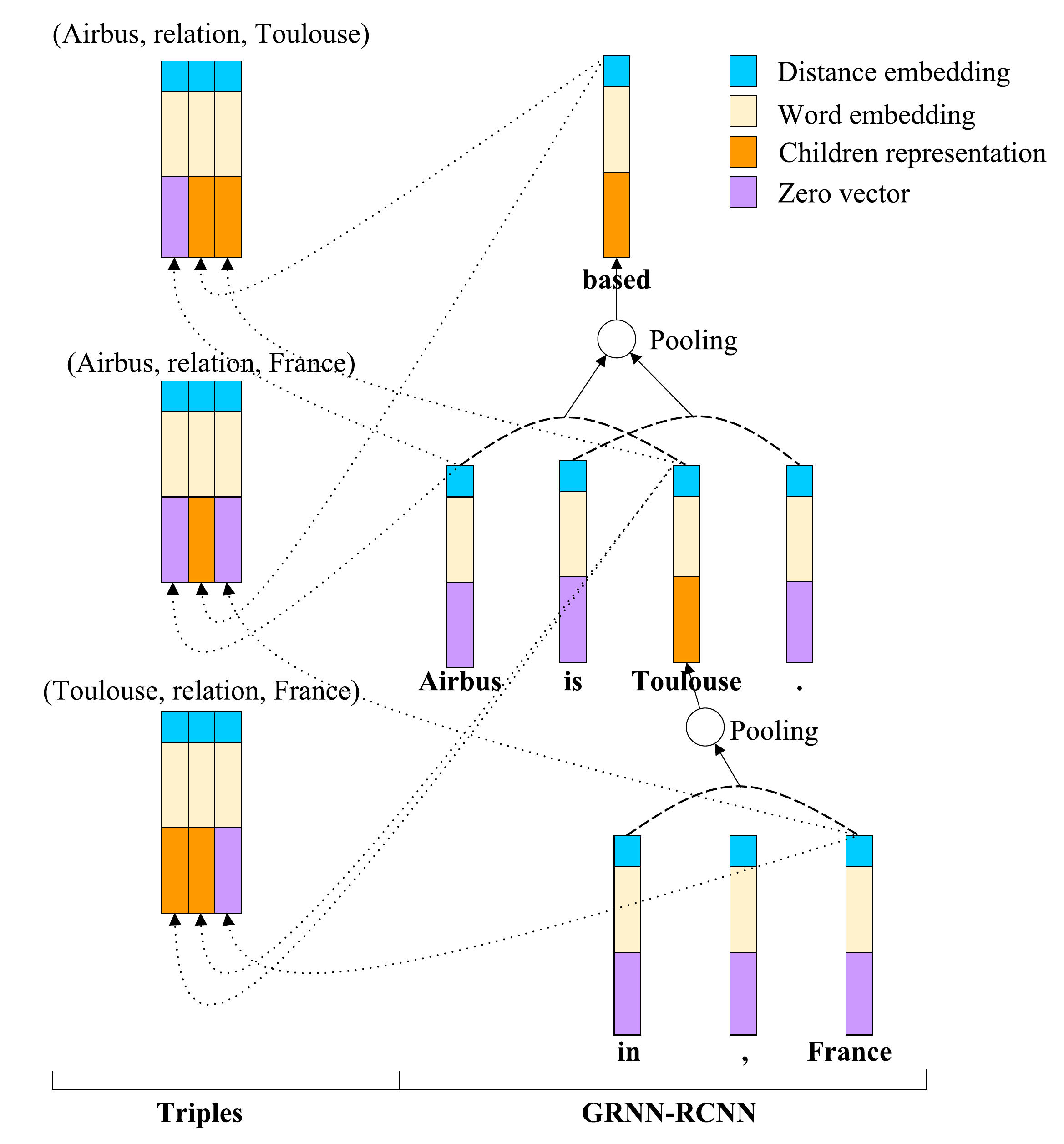}
\caption{The GRNN-RCNN extraction example of "Airbus is based in Toulouse, France."}
\label{cnn}
\end{figure}

Convolutional neural networks [\cite{lecun1998gradient}] utilize layers with convolving filters to extract local features. CNN models have been proven effective for many NLP tasks [\cite{collobert2011natural}].
\begin{align}
H^{(t)}=tanh(W*X) 
\end{align}
where $*$ denotes the 1-D convolution. $W \in \mathbb{R}^{c \times l}$ is the convolving filter where $c$ is window size and $l$ is vector dimension. As shown in Figure \ref{cnn}, the left part is the predicted triples and the right part is a graph RCNN network. The input $X \in \mathbb{R}^{P(t) \times l}$ is composed of $v_{(t,i)}$. Let $\oplus$ represent the concatenation operation.
\begin{align}
X&=[v_{(t,1)}, v_{(t,2)} ..., v_{(t,P(t)}]\\
v_{(t,i)}&=x_{t} \oplus h_{t(i)}\oplus d_{(t,i)} 
\end{align}
where the $x_{t}$ is the word embedding of the current (source) vertex. The $h_{t(i)}$ $(i=1,...P(t))$ denotes the representation of the $i$-th target vertex of source vertex $t$. 
$d_{(t,i)}$ is the distance embedding [\cite{zhu2015re}] of vertex $i$.
The distance embedding is a way to represent the relative distance between the source vertex $t$ and the $i$-th target vertex with a fixed length vector. To keep the order invariant, for the target vertices, this network uses the natural order of words in the sentence. The vertex without any target vertex consists of its word embedding and a zero vector. The output of the convolution operation is $H^{(t)}=[h_1,h_2,...,h_K]$ where $K$ is dynamic depending on the number of target vertices. Then the pooling operation captures the most informative features on rows. 
\begin{align}
h_{t}=\max\limits_{j}H_{ij}^{(t)}
\end{align}

The above neural units enhance the graph data processing through different feature extraction operations. To reduce the computational complexity, we set the maximum recursive depth to 6. The uni-directional GRNN is a top-down process, and we also built the bidirectional GRNN which can capture the features of both top-down and bottom-up directions.

\subsection{Joint Learning Networks}
\label{joint}
We employ two networks to integrate our neural units. The main difference between the two networks is the way they decode each subtask.

{\bf Joint task learning network.} As shown in Figure \ref{figarch}, the higher layer uses two decoders to jointly learn task-specific representations. We refer to it as joint task learning (JTL).

For the NER subtask, we adopt the BIOES scheme [\cite{ratinov2009design}] in a uni-directional LSTM-RNN. 
To keep more influence of the previous step this decoder also inputs the previous hidden state to the current step. 
For the RE subtask, the input of GRNN is a graph where each vertex is mapped to the contextual representation of the encoder layer.
The edge embeddings are jointly learned to control children's streams. The vertices of GRNN are mainly used to construct triple representations for the RC subtask.

\begin{figure}[!h]
\centering
\includegraphics[width=2.2in]{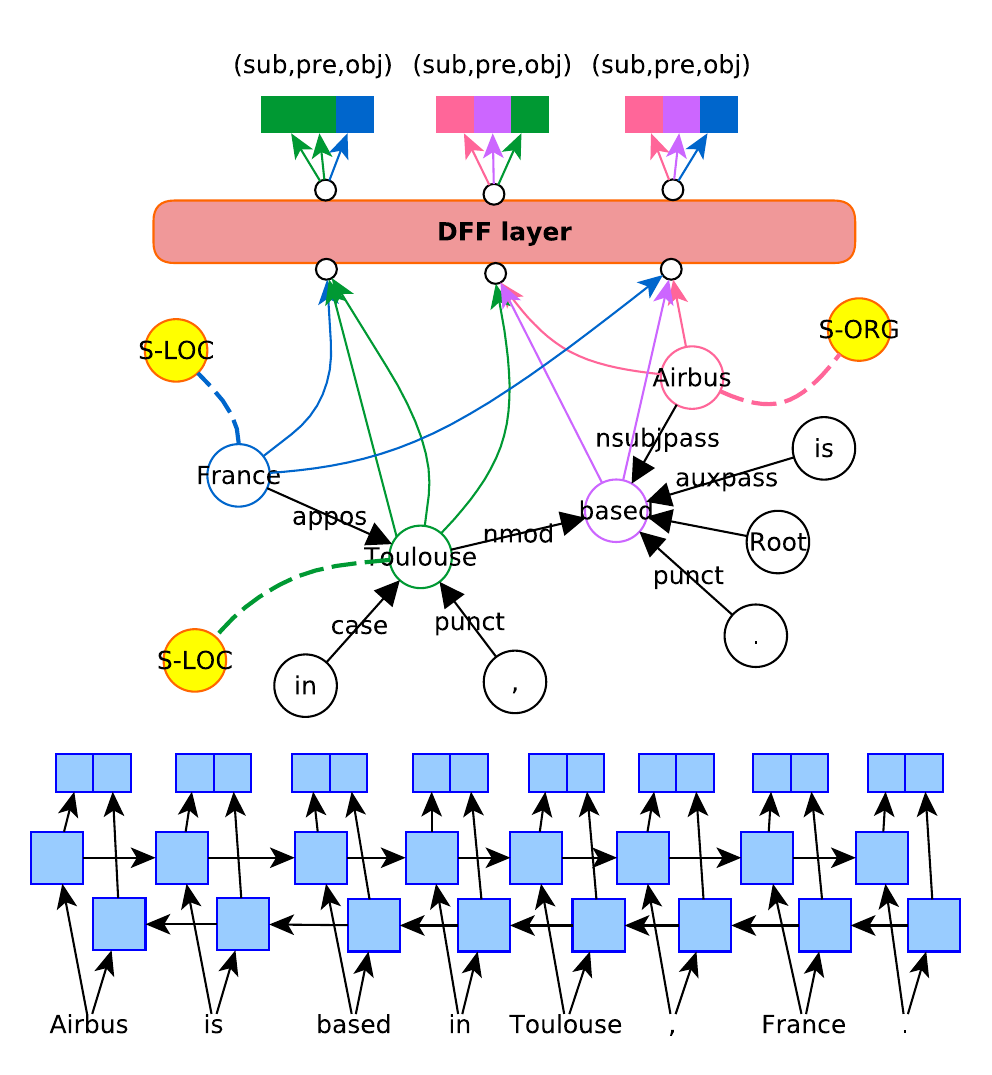}
\caption{JRL network for integrating subtasks and representations}
\label{standalone}
\end{figure}

{\bf Joint representation learning network.} 
To test whether this network can learn a unified representation [\cite{zhu2023fqp,zhu2022financial}], we simplify the network by only keeping the standalone decoder. We refer to it as the joint representation learning (JRL) network, as shown in Figure \ref{standalone}. The output of the encoder layer is input to the GRNN sub-network, and then the two subtasks directly use the representation of these vertices for prediction.

\subsection{Training Algorithm}
\label{dff}

During model training, the DFF algorithm has an inference operation that predicts entities and dynamically generates triple candidates using all possible combinations of entities. This algorithm enables the model to dynamically adapt to new patterns in the data. The downside is that there is a sort of information loss [\cite{zhu2022switchnet}] in the discrete inference process. 

This algorithm mainly contains two steps, inference, and feedback, as shown in algorithm \ref{alg.dff}. 

(1) In the inference step, this model first predicts/infers the label sequence, as shown in line 1, and uses the combination of the predicted entities as triple candidates, as shown in lines 2-3. Then, the model dynamically extracts subgraphs of multiple triples to construct the forest on the top layer, as shown in lines 5-9. This model uses the indices of $s$ and $o$ to find the LCA index as $p_w$. Then, the model will map the above indices into the GRNN to get the vector representations for a triple as shown in line 8. In lines 10-16, this model also memorizes the unexpected entity pairs that are not predicted, using the same operations of lines 5-9.

(2) In the feedback step, this model generates the relation type assumption for each predicted triple by using a single-layer neural network, as shown in line 17. Then, it compares the assumptions with the ground truth to get the feedback signal to update the model state through backpropagation. The feedback signal is generated by the loss function to calculate the gradients, as shown below.
\begin{small}
\begin{align}
\label{objective}
\max\limits_{\theta}L&=\sum\limits_{i=1}^{|\mathbb{D}|}(\log\prod\limits_{j} p_{ner}(y_{ner_{j}}^{(i)}|{\bf z}_i;\theta) + \log\prod\limits_{k} p_{re}(y_{re_{k}}^{(i)}|{\bf z}_i;\theta)) \nonumber
\end{align}
\end{small}
where the $p_{ner}$ and $p_{re}$ represent the prediction of the actual class of NER and RE subtasks respectively. $|\mathbb{D}|$ is the dataset size and $z$ denotes the input sequence. $j$ and $k$ denote the indices of the entity or relation respectively. $\theta$ is the model parameter. The final model generated by the DFF algorithm is a recursive computational graph where the parameters can be optimized jointly.

\begin{algorithm}[h]
\caption{The DFF algorithm}
\label{alg.dff}
\hspace*{\algorithmicindent} \textbf{Input}:NER decoding vectors $v$, NER labels $l_{ner}$, Triples $l_{re}$, vertex vectors $V$

\hspace*{\algorithmicindent} \textbf{Output}:The loss values of subtasks

\begin{algorithmic}[1]
 \STATE Predict the label sequence $\hat{y}_{ner}$ based on $v$
 \STATE Extract entity phrases $ens$ from the $\hat{y}_{ner}$
 \STATE Generate entity $pairs$ using binary combination of $ens$
 \STATE $\hat{y}_{rc}$=[], $l_{rc}$=[]
     \FOR{ $ j \leftarrow 0,l_{re}.length-1 $}
    \STATE $s,p,o$=$l_{re}$[$j$]
    \STATE Find the LCA index for $s$ and $o$ as $p_w$ 
    \STATE Generate $v_t$=[$v_s$,$v_{p_w}$,$v_o$] from $V$ according to indices ($s$,$p_w$,$o$), then, add $v_t$ to memory $\hat{y}_{rc}$ and add relation $p$ to $l_{rc}$ respectively
  \ENDFOR
     \FOR{$ j \leftarrow 0,pairs.length-1 $}
      \IF{$pairs$[$j$] not in $l_{re}$}
          \STATE $s,o$=$pairs$[$j$]
          \STATE Find the LCA index for $s$ and $o$ as $p_w$ 
          \STATE Generate $v_t$=[$v_s$,$v_{p_w}$,$v_o$] from $V$ according to indices ($s$,$p_w$,$o$), then, add $v_t$ to memory $\hat{y}_{rc}$ and add relation ``None'' to $l_{rc}$ respectively
        \ENDIF
     \ENDFOR
 \STATE $assump$ = predict($\hat{y}_{rc}$)
 \STATE feedback($\hat{y}_{ner}$, $l_{ner}$), feedback($assump$, $l_{rc}$)
\end{algorithmic}
\end{algorithm}

\section{Experiments}
\subsection{Experiment Setup}
\subsubsection{Dataset}
The NYT dataset is generated by aligning the Freebase relations with the news article of the 1987 $\sim$ 2007 New York Times. The training set contains 1.18M sentences with 47 entity types and 24 relation types [\cite{ren2017cotype}], i.e., ``{\it /business/company/founders}'', ``{\it /sports/sports\_team/location}'', etc. We exclude the “None” label relation, like [\cite {zheng2017joint,ren2017cotype}]. During the training process, the samples with only the ``None” label relations have little effect on the final result and we remove them. Thus, we use 66,336 training samples (about 1/3 of the training set) to reduce the training time. The test set contains 395 samples manually annotated by the author of [\cite{hoffmann2011knowledge}]. 

\subsubsection{Evaluation} 
We adopt the standard micro F1 score, recall (Rec.), and precision (Prec.) as the metrics for the NER and RE subtasks. For the final result, a correct prediction is that the extracted triple matches the ground truth including two entities, relation direction and relation type. For the NER subtask, we consider the entity type, length, and position in sentences.

\subsubsection{Hyperparameters}
The input word is projected to a 200-D pre-trained GloVe [\cite{pennington2014glove}] word embedding. The hidden state of the encoder is 300-D. The dimension of GRNN units (including the LSTM, GRU and RCNN units) is 100-D. We split the training data into 100 pieces to select better models. We first use the single-sample training to get a good model and then adopt the batch (64) training to fine-tune the model. We ran the experiments on an AMD Ryzen 5 1500X Quad-Core Processor @ 3.5GHz (Mem: 16G) and RTX 1070Ti GPUs (8G).

\subsection{Results of JTL network}
This subsection first reports the results of the 1-of-n classification form. We use Bi to represent the bidirectional modeling. The RCNN, GRU, and LSTM denote the computational units in the GRNN models.

Table \ref{results} reports the results. The first two parts are the pipeline methods and the joint extraction methods respectively. The third part is our methods where different units are implemented to augment the basic GRNN. To eliminate the influence of random factors we ran the experiments three times and take the average.

The basic GRNN (Bi-GRNN) also gets a good result, which means that the mechanism of the GRNN is effective in this task. Compared with other studies, the results of GRNN models are more balanced, while the recall and precision of other joint extraction models are not balanced enough.
This indicates that using the global optimization process allows the model to find a better balance.
The gated units improve the results since they alleviate the gradient vanishing problem. This indicates considering long-time graph dependency can help to encode rich relation representation.

\begin{table}[!h]
\centering
\caption{Results report of the joint entity and relation extraction methods on the NYT dataset}
\label{results}
\resizebox{0.45\textwidth}{!}{
\begin{tabular}{l|ccc}
\toprule
{\bf Methods}   & {\bf Prec.} & {\bf Rec.}  & {\bf F1}    \\ \midrule
FCM [\cite{mintz2009distant}] & 25.8 & 39.3 & 31.1 \\
LINE [\cite{tang2015line}] & 33.5 & 32.9 & 33.2 \\ \hline
MultiR [\cite{hoffmann2011knowledge}] & 33.8 & 32.7 & 33.3 \\
DS-Joint [\cite{li2014incremental}] & 57.4 & 25.6 & 35.4 \\
Linear-Tree [\cite{miwa2016end}]\footnote{This experiment is conducted by [\cite{wu2018indirect}] in the ReQuest system}& 37.3 & 15.4 & 23.4 \\
CoType [\cite{ren2017cotype}] & 42.3 & {\bf 51.1} & 46.3 \\
LSTM-LSTM-Bias [\cite{zheng2017joint}] & 61.5 & 41.4 & 49.5 \\ 
Transition [\cite{wang2018joint}] & {\bf 64.3} & 42.1 & 50.9 \\ \hline
{\bf Bi-GRNN} & 51.5 & 48.8  & 50.1 \\
{\bf GRNN-RCNN} & 58.3  & 45.4 & {\bf 51.0} \\ 
{\bf Bi-GRNN-GRU} & 55.5  & 48.9  & {\bf 52.0}       \\
{\bf Bi-GRNN-LSTM} & 54.9  & 50.5  & {\bf 52.6 } \\ \bottomrule
\end{tabular}}
\end{table}

\subsection{Results of JRL network}
To evaluate whether this model can obtain a unified representation, we test the JRL network of Figure \ref{standalone}. We use {\it Standalone} to represent the LSTM-based JRL network.

\subsubsection{Comparison of JTL and JRL networks }
\begin{figure}[!h]
\centering
\includegraphics[width=2.5in]{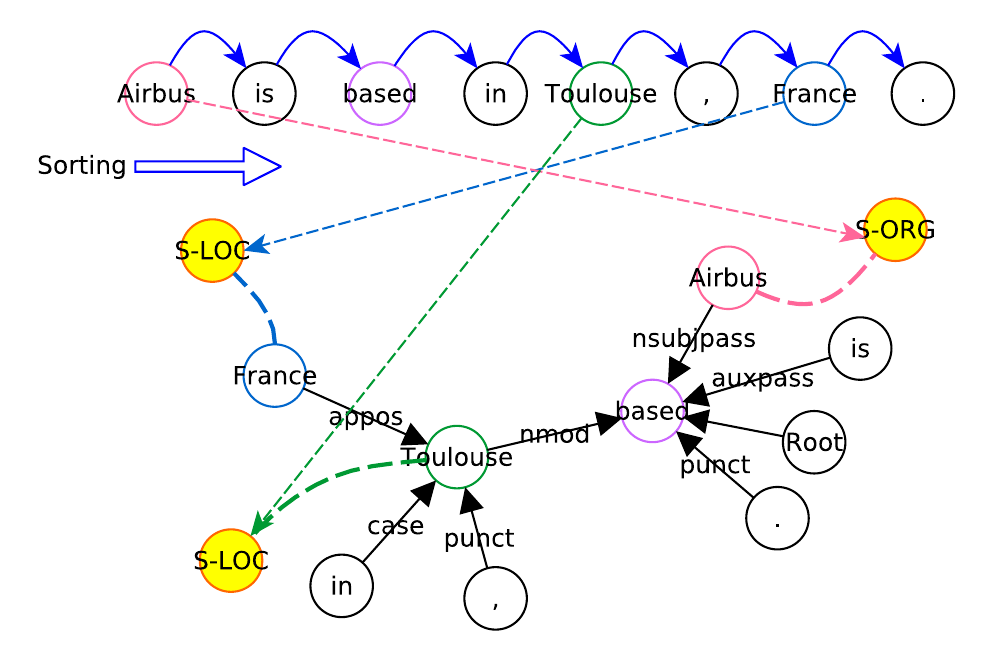}
\caption{An example of enhancing NER representation in the JRL network by sorting vertices }
\label{fig.enhance}
\end{figure}
As shown in table \ref{tab.jtrl}, the precision of the Bi-Standalone model is high but the recall decreases since we did not consider the influence of token order. To encode this factor, the Standalone model considers the bigram of vertices, more formally, $h_t^{'}=\mbox{ReLU}(W(h_{t-1}^{'}\oplus h_{t})+b)$, where $t$ is the time step and $h_t$ is the t-th vertex vector. $\oplus$ denotes the concatenation operation. $W$ and $b$ are the linear composition matrix and bias vector respectively. $h_{-1}^{'}$ is initialized to a zero vector.

This setting improves 4.40\% and 2.90\% F1 scores on the NER and RE subtasks, respectively. 
Although such a strategy is naive, by using the $W,b$ parameters, this model achieves a competitive result (52.6\% F1 score) to the LSTM decoder. This is because the NER representation is enhanced by the RE subtask by jointly updating the model. This indicates that keeping order invariant is essential.

\begin{table}[!h]
\centering
\caption{Results report where the first (1-4 rows) and the second (5-8 rows) parts are the results of JTL and JRL networks respectively}
\label{tab.jtrl}
\resizebox{0.45\textwidth}{!}{
\begin{tabular}{l|lll|lll}
\toprule
{\bf Tasks}  & \multicolumn{3}{c|}{\bf NER} & \multicolumn{3}{c}{\bf RE} \\ \hline
{\bf Methods}     & {\bf Prec.}  & {\bf Rec.}   & {\bf F1}     & {\bf Prec.}  & {\bf Rec.}  & {\bf F1}  \\ \midrule
Bi-GRNN      & 91.9  & 90.8  & 91.3  & 51.5  & 48.8 & 50.1 \\
GRNN-RCNN     & 92.1  &  91.4  & 91.7  & {\bf 58.3}  & 45.4 & 51.0 \\
Bi-GRNN-GRU  & {\bf 92.5}  & 90.9  & 91.7  & 55.5 & 48.9 & 52.0 \\
Bi-GRNN-LSTM & 92.1  &  {\bf 91.5}  & {\bf 91.8}  & 54.9 & 50.5 & {\bf 52.6 } \\  \hline
{\bf Standalone} & 84.2  & 79.8  & 81.9  & 52.9 &  41.4 & 46.4 \\ 
{\bf Bi-Standalone} & 89.3 & 85.7  & 87.4  & 53.8  & 45.9 & 49.6  \\ \hline
{\bf Standalone+sort} &91.2   &89.5    &90.3    &51.2   &48.4    &49.8   \\ 
{\bf Bi-Standalone+sort} &92.1   &{\bf 91.5}    &{\bf 91.8}  &54.5   &{\bf 50.8}  &{\bf 52.6}    \\ \bottomrule
\end{tabular}}
\end{table}
\subsubsection{Comparison of the training process }
\begin{figure}[!h]
\centering
\includegraphics[width=3.5in]{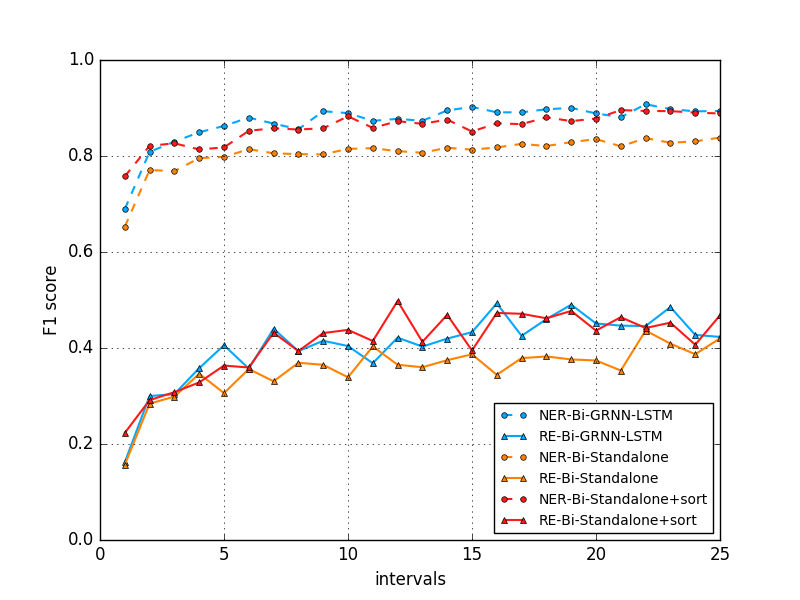}
\caption{Learning process in the first epoch (average sampling 25 intervals)}
\label{fig.epoch}
\end{figure}

To observe the training process, we split the first training epoch into 25 intervals and evaluate the models of table \ref{tab.jtrl} on the test set. As shown in Figure \ref{fig.epoch}, the JTL and JRL networks learn the NER and RE subtasks simultaneously. The learning process is effective, and some model states achieved good results. This suggests that the model selection is important in this approach.

We also observe the Bi-Standalone+sort (sorted JRL as shown in Figure \ref{fig.enhance}) achieved higher results on the RE subtask than the Bi-GRNN-LSTM. This is because the JRL network encodes more entity information in the relation representation. This experiment shows that using the same learning representation for different subtasks can help compress model parameters.

\section{Conclusion}
This paper introduces a novel approach to extracting relational triples from English textual input by leveraging graph recursive network models. The proposed methods integrate named entity recognition (NER) and relation extraction (RE) subtasks in a joint optimization process, utilizing the GRNN and the DFF training algorithm. This approach eliminates the need for designing new tag schemes and bridges the gap between subtasks by connecting the representations in both subtasks through inference operations during model training. Experimental results demonstrate the effectiveness of our model in integrating subtasks and representations. Moreover, this model can be adapted to encode subgraphs [\cite{zhu2022metaonce}] and applied in downstream applications [\cite{zhu2023metaaid,zhu2022metaaid}]. 

\end{CJK*}
\bibliographystyle{aaai}
\bibliography{reference}

\begin{thebibliography}{}

\bibitem[\protect\citeauthoryear{Chung \bgroup et al\mbox.\egroup
  }{2014}]{chung2014empirical}
Chung, J.; Gulcehre, C.; Cho, K.; and Bengio, Y.
\newblock 2014.
\newblock Empirical evaluation of gated recurrent neural networks on sequence
  modeling.
\newblock {\em arXiv preprint arXiv:1412.3555}.

\bibitem[\protect\citeauthoryear{Collobert \bgroup et al\mbox.\egroup
  }{2011}]{collobert2011natural}
Collobert, R.; Weston, J.; Bottou, L.; Karlen, M.; Kavukcuoglu, K.; and Kuksa,
  P.
\newblock 2011.
\newblock Natural language processing (almost) from scratch.
\newblock {\em Journal of Machine Learning Research} 12(Aug):2493--2537.

\bibitem[\protect\citeauthoryear{Hochreiter and
  Schmidhuber}{1997}]{hochreiter1997long}
Hochreiter, S., and Schmidhuber, J.
\newblock 1997.
\newblock Long short-term memory.
\newblock {\em Neural computation} 9(8):1735--1780.

\bibitem[\protect\citeauthoryear{Hoffmann \bgroup et al\mbox.\egroup
  }{2011}]{hoffmann2011knowledge}
Hoffmann, R.; Zhang, C.; Ling, X.; Zettlemoyer, L.; and Weld, D.~S.
\newblock 2011.
\newblock Knowledge-based weak supervision for information extraction of
  overlapping relations.
\newblock In {\em Proceedings of the ACL 2011},  541--550.
\newblock Association for Computational Linguistics.

\bibitem[\protect\citeauthoryear{Khashabi}{2013}]{khashabi2013recursive}
Khashabi, D.
\newblock 2013.
\newblock On the recursive neural networks for relation extraction and entity
  recognition.

\bibitem[\protect\citeauthoryear{LeCun \bgroup et al\mbox.\egroup
  }{1998}]{lecun1998gradient}
LeCun, Y.; Bottou, L.; Bengio, Y.; and Haffner, P.
\newblock 1998.
\newblock Gradient-based learning applied to document recognition.
\newblock {\em Proceedings of the IEEE} 86(11):2278--2324.

\bibitem[\protect\citeauthoryear{Li and Ji}{2014}]{li2014incremental}
Li, Q., and Ji, H.
\newblock 2014.
\newblock Incremental joint extraction of entity mentions and relations.
\newblock In {\em Proceedings of the ACL 2014}, volume~1,  402--412.

\bibitem[\protect\citeauthoryear{Liu \bgroup et al\mbox.\egroup
  }{2018}]{liu2018seq2rdf}
Liu, Y.; Zhang, T.; Liang, Z.; Ji, H.; and McGuinness, D.~L.
\newblock 2018.
\newblock Seq2rdf: An end-to-end application for deriving triples from natural
  language text.
\newblock {\em arXiv preprint arXiv:1807.01763}.

\bibitem[\protect\citeauthoryear{Manning \bgroup et al\mbox.\egroup
  }{2014}]{manning2014stanford}
Manning, C.; Surdeanu, M.; Bauer, J.; Finkel, J.; Bethard, S.; and McClosky, D.
\newblock 2014.
\newblock The stanford corenlp natural language processing toolkit.
\newblock In {\em Proceedings of ACL 2014},  55--60.

\bibitem[\protect\citeauthoryear{Mintz \bgroup et al\mbox.\egroup
  }{2009}]{mintz2009distant}
Mintz, M.; Bills, S.; Snow, R.; and Jurafsky, D.
\newblock 2009.
\newblock Distant supervision for relation extraction without labeled data.
\newblock In {\em Proceedings of the ACL 2009},  1003--1011.
\newblock Association for Computational Linguistics.

\bibitem[\protect\citeauthoryear{Miwa and Bansal}{2016}]{miwa2016end}
Miwa, M., and Bansal, M.
\newblock 2016.
\newblock End-to-end relation extraction using lstms on sequences and tree
  structures.
\newblock {\em arXiv preprint arXiv:1601.00770}.

\bibitem[\protect\citeauthoryear{Peng \bgroup et al\mbox.\egroup
  }{2017}]{peng2017cross}
Peng, N.; Poon, H.; Quirk, C.; Toutanova, K.; and Yih, W.-t.
\newblock 2017.
\newblock Cross-sentence n-ary relation extraction with graph lstms.
\newblock {\em arXiv preprint arXiv:1708.03743}.

\bibitem[\protect\citeauthoryear{Pennington, Socher, and
  Manning}{2014}]{pennington2014glove}
Pennington, J.; Socher, R.; and Manning, C.
\newblock 2014.
\newblock Glove: Global vectors for word representation.
\newblock In {\em Proceedings of the EMNLP 2014},  1532--1543.

\bibitem[\protect\citeauthoryear{Ratinov and Roth}{2009}]{ratinov2009design}
Ratinov, L., and Roth, D.
\newblock 2009.
\newblock Design challenges and misconceptions in named entity recognition.
\newblock In {\em Proceedings of the CoNLL 2009},  147--155.
\newblock Association for Computational Linguistics.

\bibitem[\protect\citeauthoryear{Ren \bgroup et al\mbox.\egroup
  }{2017}]{ren2017cotype}
Ren, X.; Wu, Z.; He, W.; Qu, M.; Voss, C.~R.; Ji, H.; Abdelzaher, T.~F.; and
  Han, J.
\newblock 2017.
\newblock Cotype: Joint extraction of typed entities and relations with
  knowledge bases.
\newblock In {\em Proceedings of the WWW 2017},  1015--1024.
\newblock International World Wide Web Conferences Steering Committee.

\bibitem[\protect\citeauthoryear{Socher \bgroup et al\mbox.\egroup
  }{2013}]{socher2013parsing}
Socher, R.; Bauer, J.; Manning, C.~D.; et~al.
\newblock 2013.
\newblock Parsing with compositional vector grammars.
\newblock In {\em Proceedings of the ACL 2013}, volume~1,  455--465.

\bibitem[\protect\citeauthoryear{Tang \bgroup et al\mbox.\egroup
  }{2015}]{tang2015line}
Tang, J.; Qu, M.; Wang, M.; Zhang, M.; Yan, J.; and Mei, Q.
\newblock 2015.
\newblock Line: Large-scale information network embedding.
\newblock In {\em Proceedings of the WWW 2015},  1067--1077.
\newblock International World Wide Web Conferences Steering Committee.

\bibitem[\protect\citeauthoryear{Wang \bgroup et al\mbox.\egroup
  }{2018}]{wang2018joint}
Wang, S.; Zhang, Y.; Che, W.; and Liu, T.
\newblock 2018.
\newblock Joint extraction of entities and relations based on a novel graph
  scheme.
\newblock In {\em IJCAI},  4461--4467.

\bibitem[\protect\citeauthoryear{Wu \bgroup et al\mbox.\egroup
  }{2018}]{wu2018indirect}
Wu, Z.; Ren, X.; Xu, F.~F.; Li, J.; and Han, J.
\newblock 2018.
\newblock Indirect supervision for relation extraction using question-answer
  pairs.
\newblock In {\em Proceedings of the Eleventh ACM International Conference on
  Web Search and Data Mining},  646--654.
\newblock ACM.

\bibitem[\protect\citeauthoryear{Zeng \bgroup et al\mbox.\egroup
  }{2018}]{zeng2018extracting}
Zeng, X.; Zeng, D.; He, S.; Liu, K.; and Zhao, J.
\newblock 2018.
\newblock Extracting relational facts by an end-to-end neural model with copy
  mechanism.
\newblock In {\em Proceedings of the ACL 2018}, volume~1,  506--514.

\bibitem[\protect\citeauthoryear{Zheng \bgroup et al\mbox.\egroup
  }{2017}]{zheng2017joint}
Zheng, S.; Wang, F.; Bao, H.; Hao, Y.; Zhou, P.; and Xu, B.
\newblock 2017.
\newblock Joint extraction of entities and relations based on a novel tagging
  scheme.
\newblock {\em arXiv preprint arXiv:1706.05075}.

\bibitem[\protect\citeauthoryear{Zhu \bgroup et al\mbox.\egroup
  }{2015}]{zhu2015re}
Zhu, C.; Qiu, X.; Chen, X.; and Huang, X.
\newblock 2015.
\newblock A re-ranking model for dependency parser with recursive convolutional
  neural network.
\newblock {\em arXiv preprint arXiv:1505.05667}.

\bibitem[\protect\citeauthoryear{Zhu \bgroup et al\mbox.\egroup
  }{2022}]{zhu2022switchnet}
Zhu, H.; Tiwari, P.; Zhang, Y.; Gupta, D.; Alharbi, M.; Nguyen, T.~G.; and
  Dehdashti, S.
\newblock 2022.
\newblock Switchnet: A modular neural network for adaptive relation extraction.
\newblock {\em Computers and Electrical Engineering} 104:108445.

\bibitem[\protect\citeauthoryear{Zhu}{2022a}]{zhu2022financial}
Zhu, H.
\newblock 2022a.
\newblock Financial data analysis application via multi-strategy text
  processing.
\newblock {\em arXiv preprint arXiv:2204.11394}.

\bibitem[\protect\citeauthoryear{Zhu}{2022b}]{zhu2022metaaid}
Zhu, H.
\newblock 2022b.
\newblock Metaaid: A flexible framework for developing metaverse applications
  via ai technology and human editing.
\newblock {\em arXiv preprint arXiv:2204.01614}.

\bibitem[\protect\citeauthoryear{Zhu}{2022c}]{zhu2022metaonce}
Zhu, H.
\newblock 2022c.
\newblock Metaonce: A metaverse framework based on multi-scene relations and
  entity-relation-event game.
\newblock {\em arXiv preprint arXiv:2203.10424}.

\bibitem[\protect\citeauthoryear{Zhu}{2023a}]{zhu2023fqp}
Zhu, H.
\newblock 2023a.
\newblock Fqp 2.0: Industry trend analysis via hierarchical financial data.
\newblock {\em arXiv preprint arXiv:2303.02707}.

\bibitem[\protect\citeauthoryear{Zhu}{2023b}]{zhu2023metaaid}
Zhu, H.
\newblock 2023b.
\newblock Metaaid 2.0: An extensible framework for developing metaverse
  applications via human-controllable pre-trained models.
\newblock {\em arXiv preprint arXiv:2302.13173}.

\end{thebibliography}

\end{document}